%% file: main.tex
\documentclass[10pt,twocolumn,letterpaper]{article}

\usepackage{cvpr}              
\usepackage{enumitem}
\usepackage{graphicx}
\usepackage{amsmath}
\usepackage{amssymb}
\usepackage{pifont}
\usepackage{booktabs}
\usepackage{multirow}
\usepackage{xcolor}
\usepackage{makecell}
\usepackage[pagebackref,breaklinks,colorlinks]{hyperref}

\usepackage[capitalize]{cleveref}
\crefname{section}{Sec.}{Secs.}
\Crefname{section}{Section}{Sections}
\Crefname{table}{Table}{Tables}
\crefname{table}{Tab.}{Tabs.}

\def\cvprPaperID{441} 
\def\confName{CVPR}
\def\confYear{2023}

\begin{document}

\title{Intrinsic Physical Concepts Discovery with Object-Centric Predictive Models}

\author{Qu Tang \textsuperscript{1,2$\star$} \quad Xiangyu Zhu \textsuperscript{1,2$\star$} \quad Zhen Lei \textsuperscript{1,2,3} \quad Zhaoxiang Zhang \textsuperscript{1,2,3}\thanks{$Corresponding \enspace author$. \quad $^{\star}Equal \enspace contribution$}\\
$^1$School of Artificial Intelligence, University of Chinese Academy of Sciences\\
$^2$MAIS, Institute of Automation, Chinese Academy of Sciences\\
$^3$Centre for Artificial Intelligence and Robotics, Hong Kong Institute of Science \& Innovation, \\Chinese Academy of Sciences\\
\texttt{\{tangqu2020,zhaoxiang.zhang\}@ia.ac.cn}, \texttt{\{xiangyu.zhu,zlei\}@nlpr.ia.ac.cn}
}
\maketitle

\begin{abstract}
   The ability to discover abstract physical concepts and understand how they work in the world through observing lies at the core of human intelligence. The acquisition of this ability is based on compositionally perceiving the environment in terms of objects and relations in an unsupervised manner. Recent approaches learn object-centric representations and capture visually observable concepts of objects, e.g., shape, size, and location. In this paper, we take a step forward and try to discover and represent intrinsic physical concepts such as mass and charge. We introduce the \uppercase{phy}sical \uppercase{c}oncepts \uppercase{i}nference \uppercase{ne}twork (PHYCINE), a system that infers physical concepts in different abstract levels without supervision. The key insights underlining PHYCINE are two-fold, commonsense knowledge emerges with prediction, and physical concepts of different abstract levels should be reasoned in a bottom-up fashion. Empirical evaluation demonstrates that variables inferred by our system work in accordance with the properties of the corresponding physical concepts. We also show that object representations containing the discovered physical concepts variables could help achieve better performance in causal reasoning tasks, i.e., ComPhy. 
\end{abstract}

\section{Introduction}
    \input{f1}
    Why do objects bounce off after the collision? Why do magnets attract or repel each other? Objects cover many complex physical concepts which define how they interact with the world\cite{chen2022comphy}. Humans have the ability to discover abstract concepts about how the world works through just observation. In a preserved video, some concepts are obvious in the visual appearances of objects, like location, size, velocity, etc, while some concepts are hidden in the behaviors of objects. For example, intrinsic physical concepts like mass and charge, are unobservable from static scenes and can only be discovered from the object dynamics. Objects carrying the same or opposite charge will exert a repulsive or attractive force on each other. After a collision, the lighter object will undergo larger changes in its motion compared with the massive one. How can machines learn to reveal and represent such common sense knowledge? We believe the answer lies in two key steps: decomposing the world into object-centric representations, and building a predictive model with abstract physical concepts as latent variables to handle uncertainty.  
    
    Object-centric representation learning\cite{burgess2019monet,locatello2020object,greff2019multi,tang2021object,zablotskaia2020unsupervised,singh2022simple} aims at perceiving the world in a structured manner to improve the generalization ability of intelligent systems and achieve higher-level cognition. VAE\cite{kingma2013auto}-based models, like IODINE\cite{greff2019multi} learn disentangled features that separate interpretable object properties (e.g., shape, color, location) in a common format. However, abstract concepts like mass and charge, can not be distilled by generative models since building object-level representations does not necessarily model these  higher-level-abstracted physics. CPL\cite{chen2022comphy} successfully learns high-level concepts with graph networks, but it relies on supervision signals from the ground-truth object-level concept labels. There are also several studies investigating the effectiveness of object-centric representations in learning predictive models to solve predicting and planning tasks\cite{kipf2019contrastive,hsieh2018learning,veerapaneni2020entity,tang2021object}. Nevertheless, to our knowledge, there is no work yet trying to discover and represent object-level intrinsic physical concepts in an unsupervised manner. 
    
    The idea that abstract concepts can be learned through prediction has been formulated in various ways in cognitive science, neuroscience, and AI over several decades\cite{nortmann2015primary,maus2013motion}. Intuitively, physics concepts such as velocity, mass, and charge, may emerge gradually by training the system to perform long-term predictions at the object representation level\cite{lecun2022path}. Through predictions at increasingly long-time scales, more and more complex concepts about how the world works may be acquired in a bottom-up fashion. In this paper, we focus on the main challenge: enabling the model to represent and disentangle the unfolded concepts. We follow common sense: with a neural physics engine, if the prediction of an object trajectory fails, there must be physical concepts that have not been captured.  Therefore, a latent variable that successfully models the uncertainty of prediction defines a new physical concept, and a better physical engine can be built. Following this idea, we categorize physical concepts into three levels of abstraction: extrinsic concepts, dynamic concepts, and intrinsic concepts. Firstly, the extrinsic properties (e.g., color, shape, material, size, depth, location) can be referred to as object contexts belonging to the lowest level of abstraction, and a perception module can directly encode the contexts. Secondly, the dynamic properties (e.g., velocity) in the middle level are hidden in the temporal and spatial relationships of visual features and should be inferred from short-term prediction. Thirdly, intrinsic properties like mass and charge can neither be directly observed nor inferred from short-term prediction. They can only be inferred by analyzing the way how objects exert force on each other. For example, inferring mass needs the incorporation of a collision event, and inferring charge needs to observe the change of object dynamics, which depends on a long-term observation or prediction.
    
    In this work, we build a system called PHYsical Concepts Inference NEtwork (PHYCINE). In the system, there are features arranged in a bottom-up pyramid that represents physical concepts in different abstraction levels. These features cooperatively perform reconstruction and prediction to interpret the observed world. Firstly, the object context features reconstruct the observed image with a generative model. Secondly, object dynamics features predict the next-step object contexts by learning a state transition function. Finally, the mass and charge features model the interaction between objects. PHYCINE uses a relation module to calculate pair-wise forces for all entities, and adaptively learn the variables that represent object mass and charge with proper regularization. During training, all representations are randomly initialized and iteratively refined using gradient information about the evidence lower bound (ELBO) obtained by the current estimate of the parameters. The model can be trained using only raw input video data in an end-to-end fashion. As shown in Figure 1, taking a raw video as input, PHYCINE not only extracts extrinsic object contexts (i.e., size, shape, color, location, material), but also infers more abstract concepts (i.e., object dynamics, mass, and charge). In our experiments, we demonstrate the model’s ability to discover and disentangle physical concepts, which can be used to solve downstream tasks. We evaluate the learned representation on ComPhy, a causal reasoning benchmark.
    
    Our main contributions are as follows: (i) We challenge a problem of physical concept discovery by observing raw videos and successfully discovering intrinsic concepts of velocity, mass, and charge. (ii) We introduce a framework PHYCINE, a hierarchical object-centric predictive model that infers physical concepts from low (e.g., color, shape) to high (e.g., mass, charge) abstract levels, leading to disentangled object-level physical representations. (iii) We demonstrate the effectiveness of the representation learned by PHYCINE on ComPhy.

\input{f2}
 
\section{Related Work}
    \textbf{Object-centric representation learning.} \quad Several recent lines of work learn object-centric representations from visual inputs without explicit supervision. A common framework for static scenes is returning a set of object representation vectors or slot vectors via a form of auto-encoders. MoNet\cite{burgess2019monet} firstly segment the image with an attention module, then iteratively explain away and reconstruct the scene. IODINE\cite{greff2019multi}, which is closely related to our work, uses amortized variational inference to refine the posterior parameters. Slot-Attention\cite{locatello2020object} learns a grouping strategy to decompose input features into a set of slot representations.  Further, many researchers have used architectures shown to be effective for static images to extract object-level representations from videos. ViMON\cite{weis2021benchmarking} expands MoNet to videos by adding a gated recurrent unit to aggregate information over time. Similarly, Steve\cite{singh2022simple} applies slot attention to more complex and naturalistic videos by using a transformer-based image decoder conditioned on slots.
    
    \textbf{Object-centric predictive models.} \quad Object-centric models are used to reach multi-object 3D scene prediction with or without action conditioning. PROVIDE\cite{zablotskaia2020unsupervised} fits IODINE to videos leveraging a 2D-LSTM and can be used to predict the trajectories of each object separately. ODDN\cite{tang2021object} achieves prediction with interaction by explicitly distilling object dynamics and modeling pair-wise object relations. In C-SWM\cite{kipf2019contrastive}, contrastive learning is proposed to learn a transition model in latent space. OP3\cite{veerapaneni2020entity} uses IODINE as the inference model with action conditioned, achieving object-centric prediction and planning. 

    focus on developing new methods and techniques to model a structured, dynamic environment. Specifically, they address the challenges of discovering and disentangling the rules that govern how entities interact, inducing high-level variables from low-level observables, and jointly discovering abstract representations and causal structures. In addition, these works utilize advanced machine learning techniques such as attention mechanisms and recurrent architectures to learn and represent these complex systems. 
    
    \textbf{Physical reasoning.} \quad Our work is also closely related to research on learning scene dynamics for physical and causal reasoning. Regardless of earlier benchmarks that focused on images or environments with symbolic representation, CoPhy\cite{baradel2019cophy} studies physical dynamics prediction in a counterfactual setting. CATER\cite{girdhar2019cater} introduces a synthetic video dataset for temporal reasoning associated with collocational actions; CLEVRER\cite{yi2019clevrer} incorporating dynamics modeling with compositional causal reasoning and grounding the reasoning tasks to the language domain; ComPhy\cite{chen2022comphy} studies objects’ intrinsic physical properties from objects’ interactions and how these properties affect their motions in future and counterfactual scenes to answer corresponding questions. There are lines of work that focus on developing new methods to model structured, dynamic environments\cite{alias2021neural,goyal2019recurrent,goyal2020object,ke2021systematic}. Specifically, they address the challenges of discovering and disentangling the rules that govern how entities interact, inducing high-level variables from low-level observations. While our system explicitly builds the interaction module between entities and the goal is to infer intrinsic properties by observing the dynamics.
    
\section{Method}
    
    Our goal is to learn an object-oriented and disentangled representation of physical concepts from observing the environment state. In addition, we would like to learn a predictive system that models interactions between objects under the adjustment of intrinsic physical concepts. We explicitly factorize the object representations by segmenting the latent variables $z$ into contexts $z_{ctx}$, dynamics $z_{dyn}$, mass $z_m$, and charge $z_c$ as shown in Figure \ref{fig:f2}. $z_{ctx}$ models all the extrinsic properties, including color, shape, material, location, size, etc. $z_{ctx}$ can be inferred by reconstructing the current time step image. $z_{dyn}$ mainly encodes velocity which directly determines the transition of $z_{ctx}$ to the next time step, and could be inferred by one-step prediction with a state transition function. In simple cases, $z_{ctx}$ and $z_{dyn}$ have been enough to simulate the scenes with only uniform linear motions. However, when objects exert forces on each other through different forms of interactions, such as attraction, repulsion, or collision, the object velocities would be affected by the forces, and $z_m$ and $z_c$ are required to model these interactions and update $z_{dyn}$. During training, extrinsic physical concepts can be inferred from the reconstruction of the image, and intrinsic physical concepts may emerge by enforcing the system to perform longer-term predictions. Besides, increasingly abstract levels of concepts are acquired in a hierarchical fashion, and the complete object representation is composed by the cascade of concepts of different abstract levels.  
    
    We start by introducing a generative model of multi-object representations that infers object contexts in Section \ref{sec:3.1}. Then, in Section \ref{sec:3.2}, we explain how PHYCINE infers object dynamics through short-term prediction. Finally, we introduce the interaction model and the way it helps infer mass and charge with long-term prediction in Section \ref{sec:3.3}.
    
    \subsection{Learning Object Contexts with Reconstruction }
        \label{sec:3.1}
        To achieve a systematic generalization like humans, the multi-slot framework introduced by IODINE decomposes a static image into $K$ slots, and each ideally describes an independent object or background of the input scene. The slots can capture the object’s unique contexts such as color, shape, location, and size in a disentangled style. The image $x$ is modeled with a spatial Gaussian mixture model where each slot corresponds to a single object. With a generative model, each latent vector $z_{ctx}^{(k)}$ could be decoded into a pixel-wise mean $\mu_i^{(k)}$ and a pixel-wise mask $m_i^{(k)}$ describing the assignment probability and appearance of pixel $i$ for object $k$. The generative mechanism is shared such that any ordering of the $K$ slots produces the same image. The $K$ mask logits are normalized with a softmax function to ensure that the sum of $K$ mask logits for each pixel is 1:
        \begin{equation}
        p(x|z_{ctx}) = \prod_{i=1}^D\sum_{k=1}^Km_i^{(k)}\mathcal{N}(x_i;\mu_i^{(k)},\sigma^2),
        \end{equation}
        where $\mu$ and $m$ parameterize the final spatial Gaussian mixture distribution, and $\sigma$ is the same and fixed for all $i$ and $k$. 
        
        The representation is inferred by amortized variational inference:
        \begin{equation}
        z_{ctx} \sim q_{\lambda}(z_{ctx}|x); \lambda = \left\{\mu,\sigma\right\}.
        \end{equation}
        \begin{equation}
        \lambda \leftarrow \lambda + f(z_{ctx},x, \mu, m, \nabla_\mu p(x|z_{ctx}), \nabla_m p(x|z_{ctx})).
        \end{equation}
        $q(z_{ctx}|x)$ is the posterior probability of $z_{ctx}$ given the image $x$, with the parameters of $\lambda$, $f$ is a network for parameter updating.  The posterior parameters start with an arbitrary guess, and the latent representations $z_{ctx}$ are firstly sampled from $q_\lambda$. In each refinement step, input image $x$ together with a set of variables (means $\mu$, masks $m$, mean gradient $\nabla_\mu p(x|z_{ctx})$, mask gradient $\nabla_m p(x|z_{ctx})$), which are computed from $z$ and $x$ are fed into a refinement network $f$ to obtain the additive updates for the posterior. 
    
    \subsection{Learning Object Dynamics with Short-term Prediction}
        \label{sec:3.2}
        
        Since object appearance is sufficient to conduct state image reconstruction, the object-context model can not infer dynamics. Given $z_{ctx}$, PHYCINE represents object dynamics $z_{dyn}$ by extending the idea of amortized variational inference. From the perspective of common sense, object dynamics (e.g., velocity) transit object contexts (location for rigid objects in 3D scene) in each time step. PHYCINE learns a transition function that outputs next-step contexts $z_{ctx}^{t+1}$ with $z_{ctx}^t$ and $z_{dyn}^t$ as inputs. Mathematically:
        \begin{equation}
            z_{ctx}^{t+1} = FC(z_{dyn}^t) + z_{ctx}^t.
        \end{equation}
        
        In the beginning, the predicted $z_{ctx}^{t+1}$ could be arbitrary values, and as the training goes on, the transition function is enforced to apply dynamics to location and depth other than other irrelevant concepts, e.g., shape and color. Specifically, $z_{ctx}^{t+1}$ is required to generate $x^{t+1}$, with the shared decoder as used in the reconstruction task. In order to predict the future frame image, $z_{dyn}^t$ need to be well inferred. This process could be done in one-step prediction, and could be described as:
        \begin{equation}
        z_{dyn}^t \sim q_{\lambda}(z_{dyn}^t|x^{t+1}, z_{ctx}^t).
        \end{equation}

        We can see that $z_{ctx}$ could be directly inferred by observing the image, and the discovery of $z_{dyn}$ lies in the correctness of predicting the transition of $z_{ctx}$, which makes $z_{dyn}$ of a higher abstraction level than $z_{ctx}$. However, object dynamics could be constantly changed because of intrinsic physical properties. Therefore, when higher level physical concepts get involved, $z_{dyn}^t$ could only predict $z_{ctx}^{t+1}$ correctly, since $z_{dyn}^{t+1}$ may be updated by interactions between objects, which are defined by some intrinsic concepts.     
        
    \subsection{Learning Mass and Charge with Long-term Prediction}
        \label{sec:3.3}
        In order to model the change in object dynamics, the system should consider the intrinsic concepts and disentangle the interactions from different sources. In turn, carrying out longer-term predictions endows the system with the capability of inferring the intrinsic physical concepts. To achieve this, PHYCINE explicitly designs collision force, as well as the attraction and repulsion force (i.e., the charge and force).
        
        The interaction model (Figure \ref{fig:f2} (b)) incorporates collision force and charge force to transit $z_{ctx}$ and $z_{dyn}$ from time step $t=0$ to $t=N$. By enforcing $z_m$ and $z_c$ to interpret the two kinds of forces, and keeping $z_m$ and $z_c$ constant throughout the prediction process, the model can adaptively learn the concepts of mass and charge in these two variables.
            
            \textbf{Collision.} \quad  Objects exert force on each other and change their velocities at the moment of collision. The system should capture the moment when a collision event occurs, and how the velocities of the participants change. In the model, $z_{ctx}$ and $z_{dyn}$ are used to estimate whether objects are getting close and the direction of the force. $z_{dyn}$ and $z_{m}$ are used to estimate the intensity of the force.
            
            Specifically, we first concatenate the contexts and dynamics of each object-pair and perform self-attentions to get pair-wise attentions $attn$ and force directions $\vec{F}$:
            \begin{equation}
            attn^{k_i,k_j} = W([z_{ctx}^{(k_i)};z_{dyn}^{(k_i)};z_{ctx}^{(k_j)};z_{dyn}^{(k_j)}]),
            \end{equation}
            \begin{equation}
            \vec{F}^{k_i,k_j} = W([z_{ctx}^{(k_i)};z_{dyn}^{(k_i)};z_{ctx}^{(k_j)};z_{dyn}^{(k_j)}]),
            \end{equation}
            where attention $attn^{k_i,k_j}$  represents whether or not objects $k_i$ and $k_j$ are close enough and decide whether a force is exerted, and $\vec{F}^{k_i,k_j}$ indicates the directions of the pair-wise forces. Then, we concatenate the mass and dynamics of each object-pair and estimate the force intensities:
            \begin{equation}
            |F|^{k_i,k_j} = W([z_{m}^{(k_i)};z_{dyn}^{(k_i)};z_{m}^{(k_j)};z_{dyn}^{(k_j)}]).
            \end{equation}
            With the interaction attention, force directions, and force intensities, we can get the complete force representation for each object:
            
            \begin{equation}
            F^{k_i} =  \sum_{j \neq i}{l2norm(\vec{F}*attn)}*\sum_{j \neq i}{(|F|*attn)},
            \end{equation}
            where l2 normalization ensures that $\vec{F}$ only carries the direction information.
            
            \textbf{Attraction and Repulsion.} \quad The forces of attraction and repulsion are also pair-wise interactions. Unlike the collision force, repulsion and attraction do not happen in an instant but exist all the time. 
            
            Assuming that all object pairs have a universal charge force, we estimate it from $z_{ctx}$ and $z_c$. The pair-wise product of $z_c$ can be positive or negative, or zero:
            
            \begin{equation}
            F^{k_i} =  \sum_{j \neq i}W([z_{ctx}^{(k_i)};z_{ctx}^{(k_j)}])*z_c^{(k_i)}*z_c^{(k_j)}.
            \end{equation}
            
            Forces from both modules update $z_{dyn}$ and further update $z_{ctx}$ to the next time step, and the inference of mass and charge could be formulated as:
            \begin{equation}
            z_{m}^t,z_c^t \sim q_{\lambda}(z_{m}^t,z_c^t|x^{ \leq{t+N}}, z_{ctx}^t, z_{dyn}^t).
            \end{equation}
            
        \subsection{Objective}
        In our system, Besides the initial posterior $\lambda$, the parameters of three models need to be optimized: parameters of the generative model $p_\theta$ that generate images from $z_{ctx}$;  parameters of the inference model $q_\phi$ that updates the posterior parameters using gradient information about the ELBO obtained by the current estimate of the parameters; parameters of the interaction model $r_\delta$ which roll out a sequence of $z_{ctx}^{1:N}$:
        \begin{equation}
        z_{ctx}^0, z_{dyn}^0, z_m^0, z_c^0 \sim \lambda.
        \end{equation}
        
        \begin{equation}
        z_{ctx}^{1:N}, z_{dyn}^{1:N} = r_\delta (z_{ctx}^0, z_{dyn}^0, z_m^0, z_c^0 ).
        \end{equation}
        
        We train all the parameters by gradient descent through the reconstruction and prediction objective. The training loss $\mathcal{L}$ is the negative ELBO which could be described as: 
    
        \begin{equation}
           \mathcal{L}  =  \sum_{t=0}^N(-log(p_\theta(x^t|z_{ctx}^t)) 
            +\beta KL(q_\phi(z^t|x^t,z^{<t})||p(z)))
        \end{equation}.
        The first term is the decoder negative log-likelihood, given our mixture of decoder distribution, as discussed above. The second term is the Kullback–Leibler divergence (KL) divergence of the latent posterior with the latent prior, weighted with a hyperparameter $\beta$.
    \input{t3}   
    \input{t2}   
    \input{ablation}
    \subsection{Preventing Collapse by Regularization}

        The way of predicting $z_{ctx}^{t:t+N}$ based on $z_{ctx}^{t}$ with a series of latent variables (i.e., $z_{dyx}^{t}$, $z_{m}^{t}$, and $z_{c}^{t}$) is easy to collapse when the latent variable has excessive information capacity. For example, given $z_{ctx}^{t}$, for all $z_{ctx}^{t+1}$ there exists a trivial solution that produces zero prediction loss where $z_{dyn}^{t}$ being identity with $z_{ctx}^{t+1}$. 
        
        One promising way to prevent collapse is minimizing the information content of the latent variable used in the prediction.
        We conduct this solution by pre-defining $z_{dyn}$, $z_m$, and $z_c$ as low-dimensional variables. Another purpose of this is that the low-dimensional concepts are more unambiguous and thus conducive to the discovery of the upper-level concepts. For example, 2-dimensional $z_{dyn}$ represents two orthogonal velocity directions, which remain unchanged after normalization, and this is the basis for PHYCINE to learn the concept of object mass.

        In addition to limiting the capacity of latent variables, we also limit their functions by making them work in a regularized way. As demonstrated in Figure \ref{fig:f2}:
            \begin{itemize}
            \setlength{\itemsep}{0pt}
                \item $z_{m}$ together with $z_{ctx}$ and $z_{dyn}$ model the collision.   
                \item $z_c$ and $z_{ctx}$ model the charge force.
                \item $z_{dyn}^t$ (updated after interaction) outputs the deltas between $z_{ctx}^{t}$ and $z_{ctx}^{t+1}$.
            \end{itemize}   
            
    \subsection{Training Strategy.} 
    In PHYCINE, the concept of mass is used to estimate the intensity of the force of a collision event, which requires the system to first learn the concept of collision. Besides, both collision and charge forces can change objects' velocity, it is difficult to disentangle them when joint training these two interaction mechanisms. Thus, we train PHYCINE in a bottom-up fashion with three stages. In the first stage, the model learns the concepts of context, dynamics, and collision. Specifically, we train the model on the scenes where the objects have the same mass and are uncharged. We randomly activate the collision module and enables the model to learn small attention value and dynamic change when a collision is not happening. In the second stage, we learn how mass affects the collision by training the model on scenes where objects have different masses and are uncharged with a longer-term prediction objective. In the third stage, we learn the concept of charge in the complete dataset It is worth noting that, none of the physics labels are employed to supervise the model.    
    
\section{Experiments}
    In experiments, we evaluate PHYCINE on the ComPhy dataset to illustrate the effectiveness of the learned representation. We also detailly analyze the interpretability of the discovered physical concepts.

\subsection{Dataset} 
    Compositional Physical Reasoning (ComPhy) studies objects’ intrinsic physical properties from objects’ interactions and how these properties affect their motions in future and counterfactual scenes to answer corresponding questions. Objects in ComPhy contain compositional appearance attributes like color, shape, and material. ComPhy also studies intrinsic physical concepts, mass, and charge, which can not be directly captured from objects’ static appearance. Thus, in addition to collision, there are also attraction and repulsion events in ComPhy. Accordingly, there are questions related to object concepts of mass and charge, e.g., 'Is the cube heavier than the cylinder' in Factual questions and 'If the sphere were oppositely charged, what would happen?' in counterfactual questions. ComPhy has 8,000 sets for training, 2,000 sets for validation, and 2,000 for testing. 
   
    \input{f5}  
\subsection{Video Understanding and Causal Reasoning}
    \textbf{Baselines.} \quad Approaches on causal reasoning benchmarks\cite{yi2019clevrer,chen2022comphy} can be divided into two paradigms: neural networks\cite{ding2021attention,wu2022slotformer,tang2021object,hudson2018compositional} and neuro-symbolic models\cite{chen2021grounding,ding2021dynamic,yi2019clevrer,chen2022comphy}. Neuro-symbolic models leverage various independently-learned modules and perform better than neural network baselines. ALOE is a state-of-the-art cross-modal transformer architecture that takes in object-centric visual features and word embeddings as inputs, then
    reasons the answers. ALOE performs even better than the neural-symbolic models in CLEVRER. In this work, we focus on object-centric learning and verify the learned representation with ALOE.
    
    \textbf{Results on ComPhy.} \quad We conduct experiments on ComPhy against several counterparts: CNN-LSTM, HCRN, MAC, CPL, and ALOE. Among them, CPL is a high-performance interpretable symbolic model. MAC decomposes visual question answering into multiple
    attention-based reasoning steps and predicts the answer based on the last step. ALOE is a state-of-the-art end-to-end method on CLEVRER, which serves as our baseline. 
    
    We can observe from Table \ref{tab:ret} that, video question-answering models CNN-LSTM and HCRN perform worse than the others. The reason is that models like HCRN are mostly designed for human action video, which mainly focuses on temporal modeling rather than spatial-temporal modeling of compositional physical events. CPL performs the best overall. However, explicit signals of mass and charge are used as supervision to learn the representation, pre-trained object detector is also required. MAC outperforms ALOE-IODINE on factual questions, showing the effectiveness of its compositional attention architecture. ALOE-IODINE gains the competitive edge on predictive and counterfactual questions compared with MAC. When feeding ALOE with PHYCINE features instead of IODINE, performances are boosted in all types of questions, which indicates the effectiveness of the learned representation of the intrinsic physical concepts.  
    
    We also study how a single type of physical concept (i.e., dynamics, mass, and charge) affects the results where feature dimensions are kept equal for all experiments (Table \ref{tab:ablation}). For factual questions, three types of concept representation are beneficial, and we believe it is quite reasonable. Concretely, dynamics could help answer questions like 'Are there metal cubes that enter the scene?'. mass and charge could be helpful to reason the relative weight of two objects or how they are charged. dynamics is the most effective concept in reasoning predictive questions, compared with mass or charge. This is easy to explain: It is more straightforward to estimate what will happen with an explicit representation of object velocity. In counterfactual questions, the object's weight or charge is assumed to be a counterfactual value, then the model is required to measure what subsequent events will happen. In this setting, concepts of mass and charge are crucial, which is validated in the experiments (Table \ref{tab:ret}: mass and charge improve the per-option accuracy of counterfactual questions from 63.8 to 65.5 and 66.0 respectively. More importantly, not only the representation of every single concept contributes to the results, their combination further boosts the performance.

\subsection{Video Regeneration and Prediction}
    PHYCINE infers multi-object representations with intrinsic physical concepts of one image by fitting $N$ time-steps of future frames. In other words, $z^0$ is inferred from $x^0$ to $x^N$. If the obtained representation is correct, $z^0$ should be capable of regenerating the watched video clip by firstly predicting in the latent space to get $z^1$ to $z^N$, then rendering the contexts representation to video.
    
    Evaluating how accurately the system can regenerate the given video with one set of latent representations is not enough to prove that the physical concepts are well understood. One reason is that $z^0$ may only overfit a given video without involving any physical laws (e.g., $z_{dyn}$, $z_m$, and $z_c$ encode the object location transition rather than actual intrinsic physical concepts). In order to verify whether PHYCINE fully understands how objects interact with the world with various physical concepts, we test if the system could further predict images from $x^N$ to $x^T$. Note that We only train PHYCINE to watch and predict (re-generate from the first time-step representation) images for $N$ time-steps.
    
    Figure \ref{fig:pred} shows qualitative results for the regenerated and predicted trajectories. There are two events in this dynamic scene, a collision between the sphere and the cube, and a repulsion between the sphere and the cylinder. In the beginning, the sphere and the cylinder with the same charge move toward each other. Then, the sphere collides with the stationary cube and changes its velocity. At last, the sphere and the cylinder 
    accelerate away from each other because of the repulsion force, and the cube moves with a uniform velocity. We can observe from the result that, PHYCINE could accurately capture interactions from collision force and charge force. Besides, PHYCINE can also imagine the future without further training, which shows the interpretability of our system.
 
\subsection{Interpretability and Disentanglement}
    Disentanglement is a desirable property of representations that measures how the learned features correspond to individual, interpretable attributes of objects in the scene. IODINE, as our baseline framework, encourages the emergence of object context concepts in the slots with the decoder implementing an inductive bias about the scene composition. In the design of PHYCINE, we expect the disentanglement also be maintained for the intrinsic concepts, i.e., dynamics, mass, and charge. In this section, we demonstrate the disentanglement with counterfactual prediction, which means manually adjusting the value of a certain concept and observing the corresponding effects.
    
    Figure \ref{fig:disen} shows the comparison between 5 counterfactual situations ($a \rightarrow e$) and the original inferred result, where two spheres have the same charge and are initially getting close to each other and then repelled by the charge force. There is also a collision between the blue sphere and the uncharged and stationary cube:
    \begin{enumerate}[label=(\alph*),itemsep=-2.5pt]
    \item When giving the blue sphere an opposite charge, the repulsion force turns into an attraction force and is followed by two collisions, the cube, and the red sphere, and the two spheres, which proves the correctness and interpretability of the concepts of charge and collision.
    \item When giving the uncharged cube the same charge as spheres, three objects repel and away from each other.
    \item When turning down the mass value of the cube, the collision brings more changes to its velocity. On the contrary, it gives less velocity change to the opponent. This proves the correctness and interpretability of the concept of mass.
    \item When giving the cube an initial velocity in the first dimension of $z_{dyn}$, the cube heads in one direction and avoids the collision.
    \item When giving the cube an initial velocity in the second dimension of $z_{dyn}$, the cube heads in another direction and also avoids the collision.
    \end{enumerate}
    \input{f3}
    
    The counterfactual experiments reveal the mechanisms of each dimension of the learned variables, which on the one hand consistent with the desired physical concepts, and on the other hand, are disentangled.
    
\subsection{Bottom-up Training and Regularization}
We conduct experiments to verify the importance of the bottom-up training strategy and regularization, with quantity results shown in Table \ref{tab:ablation}. 

Though regularized with the pre-defined interaction model, both mass and charge have access to directly monitor and adjust contexts, instead of through dynamics. When training step by step, the model gains a better and more accurate interpretation of the world with the accumulation of knowledge. Results indicate that the representation quality without bottom-up training decreased in the case of answering questions that related to concepts of mass and charge, i.e., Factual and Counterfactual questions.  

We get rid of regularization completely by omitting the interaction model, which means that the model needs to compress the input video into one latent variable and a simple state transition function. Results show that not only no meaningful conceptual features are learned, but also the finite-dimensional features are even worse at describing the current frame contexts than the baseline model (ctx $vs$ w/o interaction).

\section{Discussion and Conclusion}
Relations and interactions among objects are governed by physical laws which helps us interpret the configurations of objects accurately\cite{li2017visual}. In this work, we studied the configuration of intrinsic physical concepts (mass and charge) for objects, by observing their interactions. We proposed PHYCINE, a framework that infers disentangled physical concepts of different abstraction levels by fitting future physical events with amortized inference, trained using only raw input data without supervision. Based on the visualization of counterfactual predictions, it is confirmed that the inferred concepts are interpretable. The discovery of the intrinsic physical concepts helps to understand and predict physical events, and richer representations of video could be obtained, which leads to better performance in solving downstream reasoning tasks. Lastly, PHYCINE has shown the practicability of discovering and configuring intrinsic physical concepts with object-centric predictive models, it seems worth investigating further in this direction.\cite{xie2022coat}
\vspace{-7pt}
\subsubsection*{Acknowledgments}

This work was supported in part by the National Key Research \& Development Program (No. 2020YFC2003901), Chinese National Natural Science Foundation Projects \#62176256, \#62276254, \#61976229, the Youth Innovation Promotion Association CAS (\#Y2021131), and the InnoHK program.

{\small
\bibliographystyle{ieee_fullname}
\bibliography{main}
}

\clearpage

\appendix
\section{Training Configuration}
\subsection{Data Processing}
Each data point in ComPhy contains 4 reference videos and 1 target video. Within each set, the objects share the same intrinsic physical properties across all videos and each object in the target video appears at least in one of the reference videos. Reference videos have more interaction among objects, providing information for models to identify objects’ physical properties. We train PHYCINE firstly in reference videos, then fine-tune the model in target videos. Each reference video contains 50 frames, and we sample images every 4 frames. Data balance is also important to promote the convergence of the relation model. Following the training strategy in Section 3.6, we balance the data by the object's physical properties in the scene. In detail, we categorize the reference videos into 5 classes as shown in table \ref{tab:data}, and the number of videos in each category is similar. During training, sub-dataset 1 is firstly used to learn concepts of object contexts, object dynamics, and collision, then we add sub-dataset 2 to the training data to learn the concept of mass, and finally, we add the reset data to further learn the concept of charge. After all the concepts have been trained, we fine-tune the model in target videos. 
\input{data}

\subsection{Hyper-parameters and training}
We set the length of each video clip $N$ to be 6, which means the inferred scene representation for the first frame should be able to predict the rest five frames. 
For the other parameters, we Generally follow the setting of IODINE. We initialize the parameters of the posterior $\lambda$ by sampling from $U( -0.5,0.5)$. We use a latent dimensionality of 16 as ALOE which makes dim($\lambda$) = 32 and downscale the image from $320\times 480$ into 64$\times$96. The variance of the likelihood is set to $\sigma$ = 0.3 in all experiments. We keep the default number of iterative refinements at $R$ = 5 and use $K$ = 8 for both training and testing. We set $\beta$ = 100.0 for all experiments. We train our models on 8 GeForce RTX 3090 GPUs, which takes approximately 4 days per model. We use ADAM for all experiments, with a learning rate of 0.0003.

\input{f6}

\begin{figure*}[!h]
  \centering
  \includegraphics[width = 0.75\textwidth]{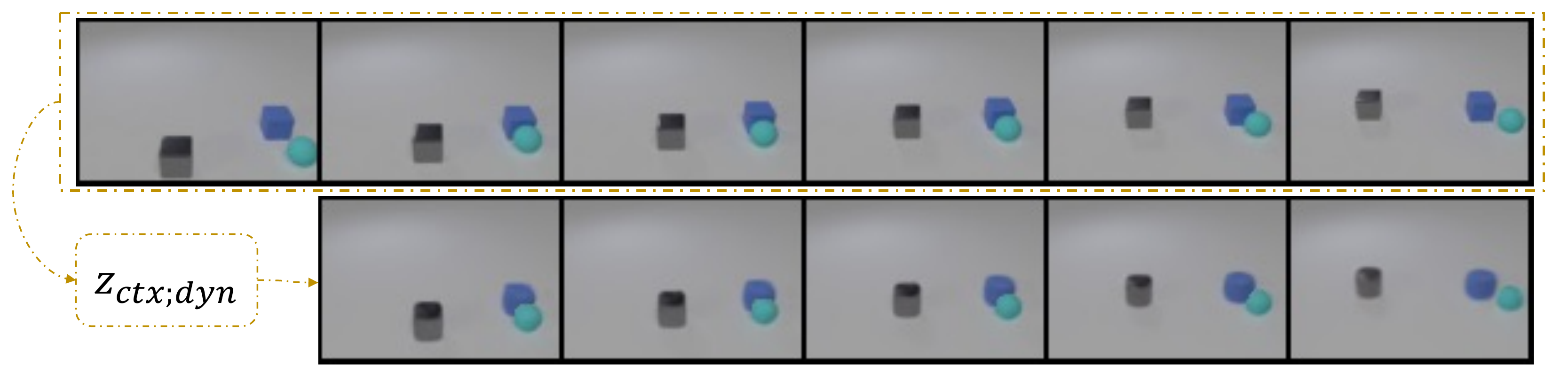}
  \vspace{-0.8em}
   \caption{Prediction with higher dimension dynamics.}
   \vspace{-2.1em}
   \label{fig:reg_dim}
\end{figure*}

\section{Necessity of regularization} 
We discuss the necessity of bottom-up training and variable content reduction with qualitative results. In Figure \ref{fig:bu}, one collision event occurs between the cylinder and the sphere. From the regeneration results, we can tell that PHYCINE without bottom-up training can not capture the collision but records the trajectories of objects. In addition, for the counterfactual setting "if the sphere is heavier", PHYCINE predicts a likely result: trajectories of both the cylinder and the sphere are affected because of this setting, and the first two frames are not affected because the collision does not happen yet. By contrast, the result of PHYCINE without bottom-up training shows that the model didn't really learn the concept of "mass", because only the sphere is affected.

In order to demonstrate the necessity of reducing variable contents, we conduct an experiment with higher dimension dynamic variables and without interaction modeling (with an FC layer only), Figure. \ref{fig:reg_dim} shows that the model can roll out videos with objects interacting even without the interaction model, suggesting that the higher-dimensional $dyn$ learns redundant information (such as change of velocities directions) that should not be entangled and the model collapse.

\section{Visualizations and numbers of intermediate variables}

\begin{figure*}[!h]
  \centering
  \includegraphics[width = 0.75\textwidth]{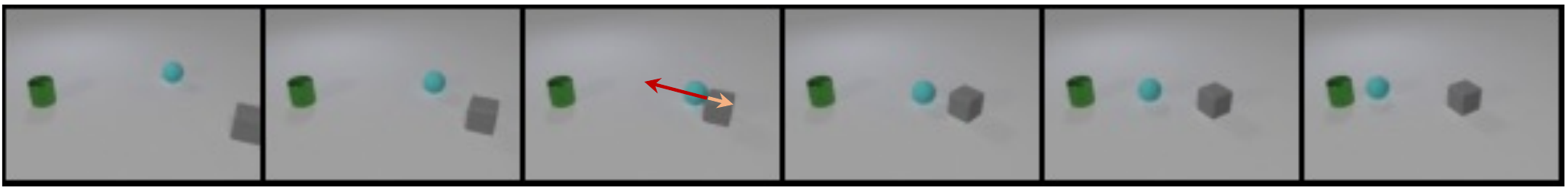}
  \vspace{-0.9em}
   \caption{Example for intermediate variable visualization.}
   \vspace{-1.2em}
   \label{fig:1}
\end{figure*}

We visualize the intermediate variables from the force modeling process of the example in Figure. \ref{fig:1} and Table. \ref{tab:viz}. In frame 3, the two objects were subjected to forces of opposite directions, the force intensities are determined by both dynamics and mass. Their dynamics are changed consequently in frame 5.
\vspace{-0.1em}

\vspace{-0.1em}
\begin{table}[h]
	\centering
	\footnotesize
	\tabcolsep=5.35pt
	\begin{tabular}{c*{8}{|c}}
		\hline
		F &$\color{cyan}dyn$ &$\color{gray}dyn$ &$\color{cyan}f_d$  &$\color{gray}f_d$ &$\color{cyan}f_i$&$\color{gray}f_i$ &$\color{cyan}m$ &$\color{gray}m$ \\
		\hline
		1   & \makecell[c]{-0.74\\-1.65} &\makecell[c]{2.21\\ -0.02}  &\makecell[c]{0.92\\-0.39} &\makecell[c]{0.83\\ -0.56} &0.01&0.01&-4.83&-0.15   \\
		\hline
		3 &\makecell[c]{-0.66\\-1.64} &\makecell[c]{2.20\\-0.04} &\makecell[c]{0.99\\0.16} &\makecell[c]{-0.99\\ -0.12} &3.67&0.87&-4.83&-0.15   \\
		\hline
        5  &\makecell[c]{3.12\\ -1.03}  & \makecell[c]{1.25\\ -0.17} &\makecell[c]{-0.26\\ 0.96}  & \makecell[c]{-0.96\\ -0.27}&0.13&0.03&-4.83&-0.15  \\
		\hline
	\end{tabular}%
	\vspace{-0.3mm}
	\caption{Variable numbers visualization. Frame (F) 1, 3, and 5 are sampled, with $f_d$ and $f_i$ indicating the force direction and the force intensity applied respectively.}
	\label{tab:viz}
\end{table}

\end{document}

%% file: f1.tex
\begin{figure}[!ht]
    \centering
    \includegraphics[width = 0.5\textwidth]{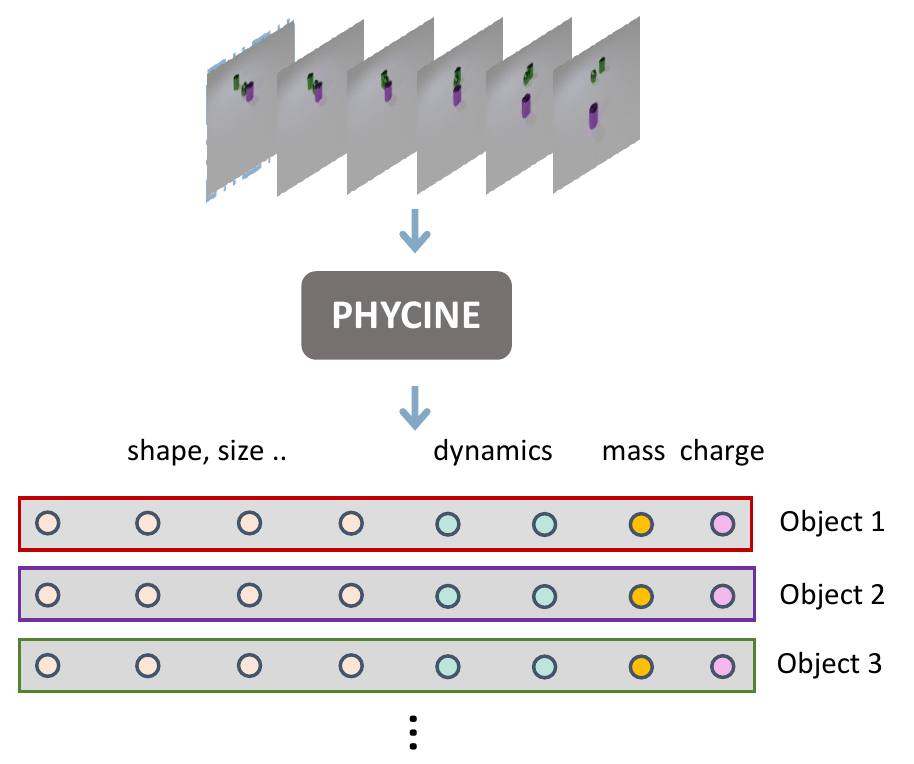}
     \vspace{-10pt}
    \caption {Given a video from the ComPhy\cite{chen2022comphy} dataset, PHYCINE decomposes the scene into multi-object representations that contain physical concepts of different abstraction levels, including visual attributes, dynamics, mass, and charge. (shown in different colors).}
    \label{fig:f1}
    \vspace{-18pt}
\end{figure}

%% file: f2.tex
\begin{figure*}[!ht]
    \centering
    \includegraphics[width = 1.0\textwidth]{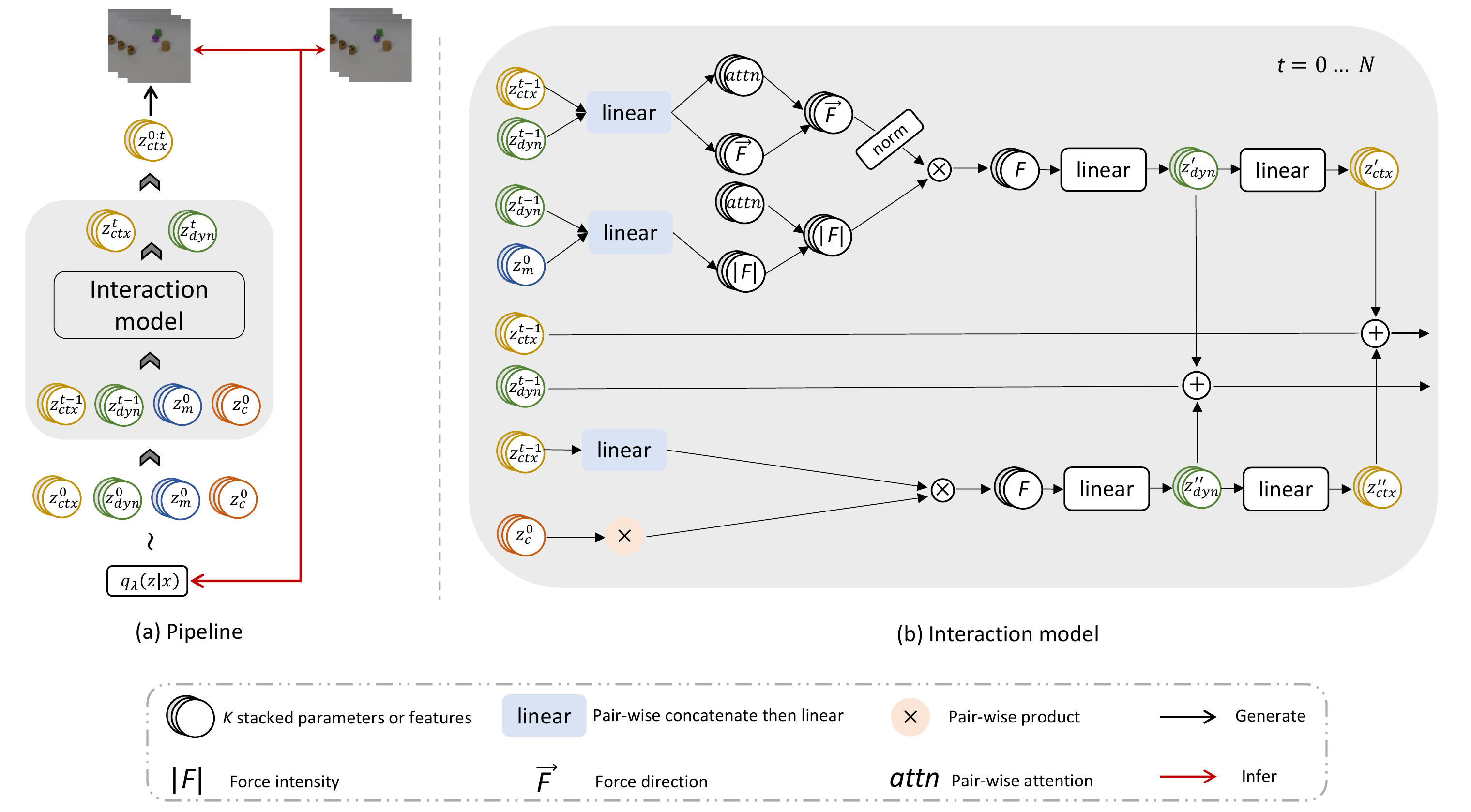}
    \vspace{-8pt}
    \caption {(a). Schematic of PHYCINE. Given a video with images of $x^0 \cdot x^t$, we aim to get four physical embeddings of each object, including context $z_{ctx}$, dynamics $z_{dyn}$, mass $z_m$ and charge $z_c$. Firstly, representations of the time-step 0 are sampled from an arbitrary posterior distribution. Then, $z_{ctx}^0$ is required to reconstruct image $x^0$, and the interaction model takes as inputs $z_{ctx}^0$, $z_{dyn}^0$, $z_{m}^0$, and $z_{c}^0$ to iteratively output $z_{ctx}^t$. In this way, a video could be rendered. The gradient information from image generation is used to iteratively refine the posterior, which could finally accurately represent the physical concepts of all present objects. (b). Interaction model that models collision force and charge force with all the identical nodes pointing to one shared tensor. }
    \label{fig:f2}
    \vspace{-12pt}
\end{figure*}

%% file: t3.tex
\begin{table}[t!]
    \tabcolsep=3pt
	\centering
    \caption{Dimensions of physics latent codes, and their abstraction levels and inference conditions among different concepts.}
	\begin{tabular}{cccc}
	\hline
		Concept & dim & abstraction &inference \\\hline
		Contexts & 12 & low & image reconstruction\\ 
	    Dynamics & 2  & middle & short-term prediction\\
	    Mass & 1 & high&long-term prediction\\
	    Charge& 1 & high& long-term prediction\\\hline
	\end{tabular}%
	\vspace{-10pt}
	\label{tab:dim}%
\end{table}%

%% file: t2.tex
\begin{table*}[t!]
    \tabcolsep=0pt
	\centering
    \caption{Question-answering accuracy of visual reasoning models on COMPHY. We report
            per-option and per-question accuracies for each sub-task. CPL uses a supervised object detector to obtain object contexts, such as Faster/Mask
            R-CNN\cite{ren2015faster,he2017mask}, then learns representations of mass and charge with object-centric labels as supervision. We omit the self-supervised learning strategy of ALOE for implementation simplicity.}
	\resizebox{0.95\textwidth}{!}{
	\begin{tabular}{lcccccc}
	\hline
		Model & learned physics& Factual & \multicolumn{2}{c}{Predictive} & \multicolumn{2}{c}{Counterfactual} \\
		& & & per opt. & per ques. & per opt. & per ques. \\ \hline
		CNN-LSTM\cite{antol2015vqa} &-& 46.6 &  59.5 & 29.8 & 58.6 & 14.6  \\ 
	    HCRN\cite{le2020hierarchical} & -&47.3 &  62.7 & 32.7 & 58.6 & 14.2  \\
	    MAC\cite{hudson2018compositional} & -&\textbf{68.6} &  60.2 & 32.2 & 60.2 & 16.0 \\
	    $CPL^{phy-label}$\cite{chen2022comphy} &ctx+m+c& {\color{gray} 80.5} & {\color{gray} 75.3} & {\color{gray} 56.4} & {\color{gray} 68.3} & {\color{gray} 29.1} \\
	    ALOE-IODINE\cite{ding2021attention,greff2019multi}  & ctx & 54.4 & 60.7 & 29.1 &63.8 &20.9  \\\hline
	    ALOE-PHYCINE &ctx+dyn+m+c&  \hspace{0.87cm} 59.1 ({\color{gray}+4.7}) & \hspace{0.9cm} \textbf{63.3} ({\color{gray}+2.6}) & \hspace{0.9cm} \textbf{33.4} ({\color{gray}+4.3}) & \hspace{0.95cm}\textbf{68.2} ({\color{gray}+4.4}) &\hspace{0.95cm}\textbf{25.2} ({\color{gray}+4.3})\\\hline
	    
	\end{tabular}%
	}
	\vspace{-5pt}
	\label{tab:ret}%
\end{table*}%

%% file: ablation.tex
\begin{table}[t!]
    \tabcolsep=3pt
	\centering
    \caption{Ablation experiments. We study the contribution of each concept representation and the importance of the bottom-to-up training strategy and interaction modeling. We adjust the contexts dimension to ensure that all compared features are the same dimension.}
	\resizebox{0.48\textwidth}{!}{
	\begin{tabular}{lccccc}
	\hline
		 & Factual & \multicolumn{2}{c}{Predictive} & \multicolumn{2}{c}{Counterfactual} \\
		&  & per opt. & per ques. & per opt. & per ques. \\ \hline
	    ctx& 54.4 & 60.7 & 29.1 &63.8 &20.9  \\
	    ctx + m & 55.2 & 62.8 & 31.5 &65.5 &22.6  \\
	    ctx + c  & 55.7 & 61.8 & 31.3 &66.0 &22.1  \\
	    ctx + dyn & 55.5 & 63.1 & 33.2 &64.4 &21.5 \\
	    ctx+dyn+m+c  & \textbf{59.1} & \textbf{63.8} & \textbf{34.2} & \textbf{68.2} &\textbf{25.2} \\\hline
	    w/o b2u training & 58.0 & 63.4& 33.6 & 67.0 &22.8 \\
	    w/o interaction & 47.3 & 57.8 & 25.5 & 63.9 & 20.1 \\\hline
	    
	\end{tabular}%
	}
	\vspace{-10pt}
	\label{tab:ablation}%
\end{table}%

%% file: f5.tex
\begin{figure*}[!t]
    \centering
    \includegraphics[width = 1.0\textwidth]{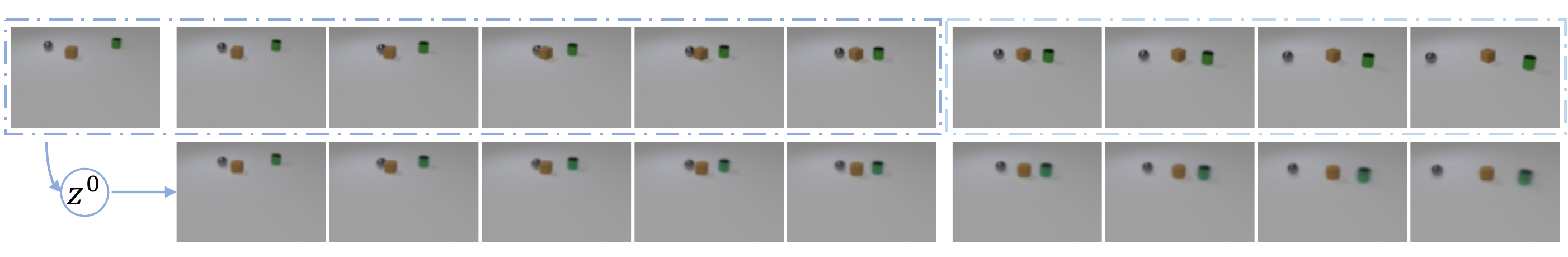}
    \caption{Example result of PHYCINE inferring $z^0$ given 6 frames, then regenerating the video and imagining the future (shown with 4 more frames).}
    \label{fig:pred}
    \vspace{-15pt}
\end{figure*}

%% file: f3.tex
\begin{figure}[!t]
    \centering
    \includegraphics[width = 0.47\textwidth]{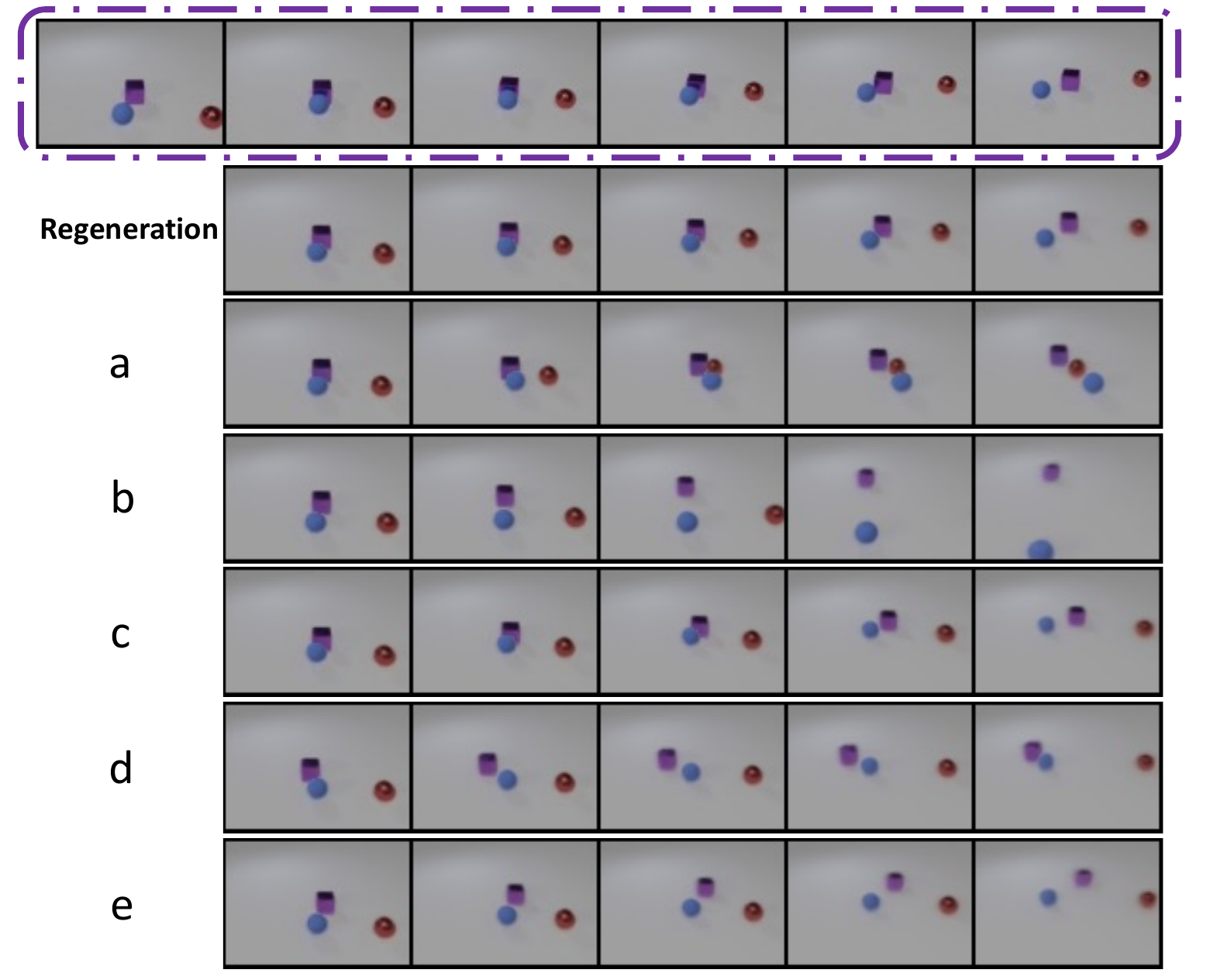}
    \caption{Disentanglement of the discovered intrinsic physical concepts, i.e., dynamics, mass, and charge, with counterfactual prediction. }
    \label{fig:disen}
    \vspace{-15pt}
\end{figure}

%% file: data.tex
\begin{table}[h!]
    \tabcolsep=3pt
	\centering
    \caption{Categorized training dataset.}
	\begin{tabular}{cccc}
	\hline
		sub-dataset & charge & collision & identical-mass \\\hline
		1 & \ding{56} & \ding{52} & \ding{52}\\ 
	    2 & \ding{56}  & \ding{52} & \ding{56}\\
	    3 & \ding{52} & \ding{56} &  \\
	    4& \ding{52}  & \ding{52}& \ding{52}\\
	    5& \ding{52}  & \ding{52} & \ding{56}\\\hline
	\end{tabular}%
	\vspace{-10pt}
	\label{tab:data}%
\end{table}%

%% file: f6.tex
\begin{figure*}[!ht]
    \centering
    \includegraphics[width = 0.9\textwidth]{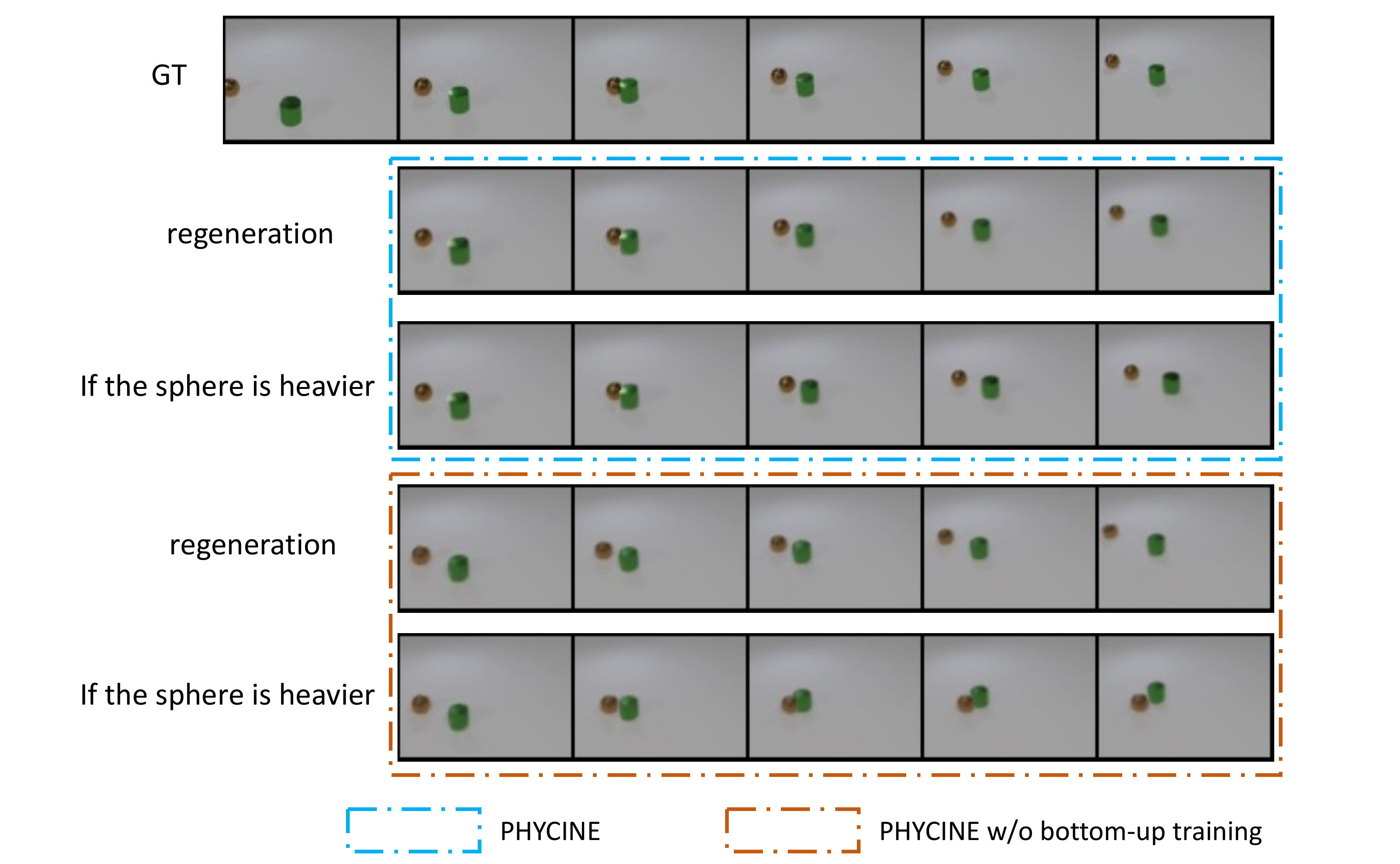}
    \caption{Visualization comparison between PYHCINE and PHYCINE without bottom-up training. Both regeneration and counterfactual results are shown.}
    \label{fig:bu}
\end{figure*}